\documentclass[letterpaper, 10 pt, conference]{ieeeconf}  

\IEEEoverridecommandlockouts                            
\usepackage[binary-units]{siunitx}
\usepackage{glossaries}
\newacronym{fmcw}{FMCW}{Frequency-Modulated Continuous-Wave}
\newacronym{vae}{VAE}{Variational Autoencoder}

\usepackage[labelfont=bf]{caption}
\usepackage{subcaption}
\usepackage{paralist}

\usepackage[dvipsnames]{xcolor}

\usepackage{tabularx,booktabs}
\newcolumntype{C}{>{\centering\arraybackslash}X} 

\usepackage{booktabs}
\usepackage{paralist}
\usepackage{textcomp}
\usepackage{algorithm}
\usepackage[labelfont=bf]{caption}
\usepackage{subcaption}
\usepackage{cite}
\usepackage{amsmath,amssymb,amsfonts}
\usepackage{algorithmic}
\usepackage{graphicx}
\usepackage{textcomp}
\usepackage{algorithm}
\usepackage{algorithmic}
\def\BibTeX{{\rm B\kern-.05em{\sc i\kern-.025em b}\kern-.08em
T\kern-.1667em\lower.7ex\hbox{E}\kern-.125emX}}

\usepackage{hyperref}
\usepackage[capitalize]{cleveref}
\crefname{section}{Sec.}{Secs.}
\Crefname{section}{Section}{Sections}
\Crefname{table}{Table}{Tables}
\crefname{table}{Tab.}{Tabs.}

\usepackage{cuted}

\renewcommand{\baselinestretch}{0.985}

\begin{document}

\title{\Large \bf Off the Radar: Uncertainty-Aware Radar Place Recognition with Introspective Querying and Map Maintenance \\
}

\author{
Jianhao Yuan, Paul Newman, Matthew Gadd\\
Mobile Robotics Group, Department of Engineering Science, University of Oxford, UK\\
\texttt{jianhao.yuan@oxfordrobotics.institute},~
\texttt{\{pnewman,mattgadd\}@robots.ox.ac.uk}
}

\maketitle

\begin{strip}
\vspace{-15mm}
\centering
\includegraphics[width=\textwidth]{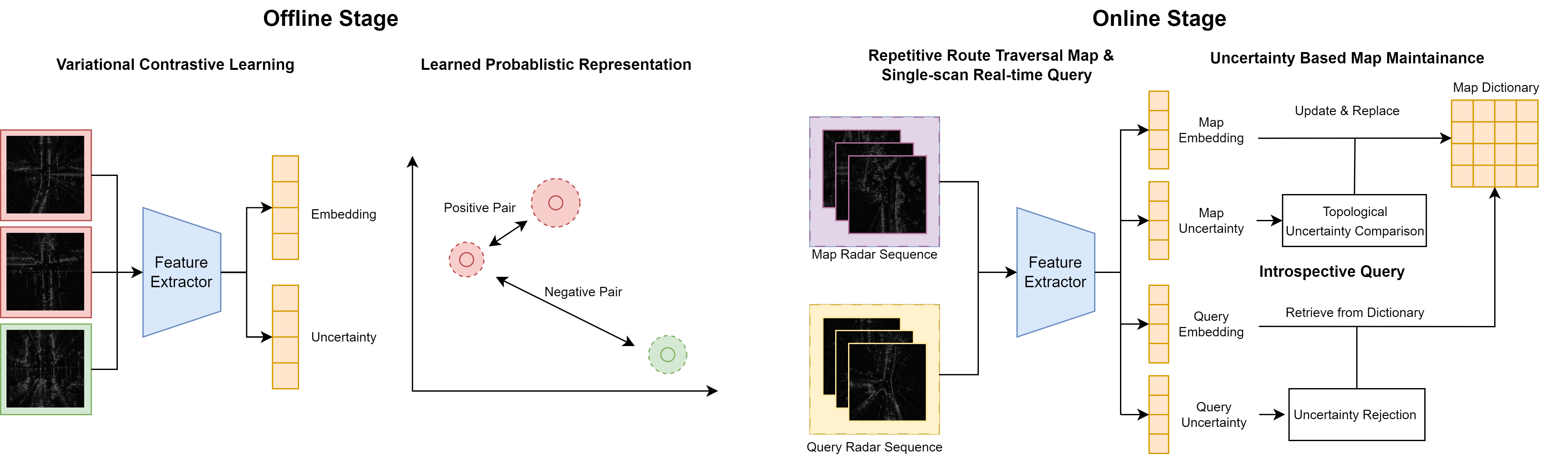}
\captionof{figure}{ {\bf Overview of system pipeline.} 
In the offline stage, we use a variational contrastive learning framework (see \cref{Sec:VarCon} for full details) to learn a hidden embedding space with estimated uncertainty such that radar scans from similar topological locations are close to each other, and vice versa. In the online stage, we develop two uncertainty-based mechanisms to handle sequentially collected radar scans for inference and map construction. For repetitive traversal of the same routes, we actively maintain an integrated map dictionary by replacing highly uncertain scans with more certain ones. For query scans with low uncertainty, we retrieve the matching map scans from the dictionary based on metric space distance. In contrast, we reject prediction for high-uncertainty scans. \vspace{-3mm}}
\label{fig:intro}
\end{strip}

\begin{abstract}
Localisation with \gls{fmcw} radar has gained increasing interest due to its inherent resistance to challenging environments.
However, complex artefacts of the radar measurement process require appropriate uncertainty estimation -- to ensure the safe and reliable application of this promising sensor modality.
In this work, we propose a multi-session map management system which constructs the ``best'' maps for further localisation based on learned variance properties in an embedding space.  
Using the same variance properties, we also propose a new way to introspectively reject localisation queries that are likely to be incorrect.
For this, we apply robust noise-aware metric learning, which both \textit{leverages} the short-timescale variability of radar data along a driven path (for data augmentation) and \textit{predicts} the downstream uncertainty in metric-space-based place recognition.
We prove the effectiveness of our method over extensive cross-validated tests of the \textit{Oxford Radar RobotCar} and \textit{MulRan} dataset.
In this, we outperform the current state-of-the-art in radar place recognition and other uncertainty-aware methods when using only single nearest-neighbour queries.
We also show consistent performance \textit{increases} when rejecting queries based on uncertainty over a difficult test environment, which we did not observe for a competing uncertainty-aware place recognition system.
\end{abstract}
\begin{keywords}
Radar, Place Recognition, Deep Learning, Uncertainty Estimation, Autonomous Vehicles, Robotics
\end{keywords}

\section{Introduction}
Place recognition and localisation are important tasks in the field of robotics and autonomous systems, as they enable a system to understand and navigate its environment.
Traditional vision-based methods for place recognition are often vulnerable to changes in environmental conditions such as lighting, weather, and occlusion, leading to performance degradation \cite{domainbed}.
To address this issue, there has been increasing interest in the use of \gls{fmcw} radar as a robust sensor substitute for such adversarial environments.
Existing works have demonstrated the effectiveness of \gls{fmcw} radar place recognition with hand-crafted \cite{checchinscan, cen2019radar,hong2020radarslam, scancontext, gskim-2020-mulran} and learning-based feature extraction approaches \cite{suaftescu2020kidnapped, gadd2021icar,komorowski2021large, gadd2020look, wang2021radarloc,demartini2020kradar,cait2022autoplace, gadd2021icraws,yin2021radar}. 
Despite the success of existing works, the deployment of these methods in safety-critical applications such as autonomous driving is still limited by the lack of calibrated uncertainty estimation.
While there have been works \cite{gal2016dropout,cai2022stun, warburg2021bayesian, shi2019probabilistic} that address the problem in similar areas such as image retrieval or visual place recognition, which leverage both Bayesian and learning-based approaches, there is no previous work specifically targeting uncertainty estimation for radar place recognition. 


In this area, it is important to consider: (1) safety requires uncertainty estimation to be well-calibrated to false positive rate in order to enable \textit{introspective} rejection; (2) real-time deployment requires \textit{fast} single-scan uncertainty-based inference capability; (3) repetitive route traversal in long-term autonomy requires online \textit{continual} map maintenance. 

While the \gls{vae} \cite{kingma2013auto} is usually used for generative tasks, its probabilistic latent space can serve as an effective metric space representation for place recognition \cite{vaepr1, vaepr2, rey2021contentbased} and allow prior assumption on the data noise distribution, which also gives a normalized aleatoric uncertainty estimation.

Thus, in this paper, to achieve reliable and safe deployment of \gls{fmcw} radar in autonomous driving, we leverage a variational contrastive learning framework and propose a unified uncertainty-aware radar place recognition method as shown in \cref{fig:intro}.
Our key contributions are as follows:
\begin{enumerate}
\item Uncertainty-aware metric learning framework for radar place recognition.
\item Introspective query mechanism based on false positive calibrated uncertainty estimation for real-time autonomous driving deployment. 
\item Online recursive map maintenance and improvement mechanism for repeated traversals of changing environments in long-term autonomy.
\end{enumerate}
In doing so we outperform the previous radar place recognition state-of-the-art~\cite{gadd2021icar} and show that our learned uncertainty is more suitable for query rejection than a previous approach~\cite{cai2022stun} developed for vision but tested here in radar.

\section{Related Work}
\subsection{Radar Place Recognition}
Place recognition can be viewed as a query and retrieval process, where a map is constructed with a dictionary of radar scans.
For every new query scan, the algorithm needs to retrieve a scan from the dictionary such that samples from a similar topological location are close to each other, and vice versa. Traditional hand-crafted feature extraction methods, such as correlative scan matching \cite{checchinscan}, graph matching \cite{cen2019radar, hong2020radarslam}, and scan context descriptor \cite{gskim-2020-mulran}, have demonstrated the effectiveness of using radar perception for place recognition and localisation. Recently, learning-based methods have shown impressive performance. Supervised metric learning methods \cite{suaftescu2020kidnapped,komorowski2021large,demartini2020kradar,gadd2020look,wang2021radarloc,cait2022autoplace}, which exploit rotational invariance and spatial-temporal consistency in radar scans, have demonstrated remarkable performance. Then, Gadd \textit{et al}~\cite{gadd2021icar} achieved comparable performance to supervised methods using contrastive learning. Moreover, multi-modal methods incorporating additional modalities, such as LIDAR points, through direct joint learning \cite{yin2021radar} or point-cloud registration \cite{8967633}, are also explored to aid radar-based recognition.
Among numerous metric learning methods, the \gls{vae} structure is extensively explored. It uses variational inference to approximate the posterior distribution of latent representation and thus gives a probabilistic estimation on embedding with consideration of data noise. Lin \textit{et al}~\cite{Lin2018dvml} use \gls{vae} in metric learning for the visual classification task. Burnett \textit{et al}~\cite{burnett2021radar} use a VAE-like structure with factor graph optimisation for radar odometry.

Inspired by previous works of the variational metric learning approach of Lin \textit{et al}~\cite{Lin2018dvml} and the contrastive learning approach of Gadd \textit{et al}~\cite{gadd2021icar}, we form the basis of our representation learning framework in a contrastive manner with a VAE-like structure. 





\subsection{Uncertainty Estimation}
Uncertainty estimation aims to quantify the model prediction confidence, which is a crucial component in the safety-critical application for the model to introspectively reject low confidence predictions. One line of work in similar areas, such as image retrieval and place recognition, uses the Bayesian approach such as Monte Carlo Dropout \cite{gal2016dropout} to estimate the epistemic uncertainty, which, however, usually is very computationally expensive and hinders the deployment in real-time autonomous driving. Another line of work directly learns the uncertainty from data to estimate the aleatoric uncertainty. However, previous approaches such as PTL \cite{shi2019probabilistic}, BPE \cite{warburg2021bayesian}, and TAH \cite{taha2019unsupervised} are restricted to specific pairwise or triplet loss, which lacks flexibility in offline system design. Later STUN \cite{cai2022stun} uses a knowledge distillation framework to break such constraints that adapt to arbitrary metric learning loss functions. 
Moreover,
STUN \cite{cai2022stun} use a dynamic binning strategy that requires a large batch of test samples to determine a meaningful uncertainty rejection threshold at inference time. Here, only the relative uncertainty within a batch can be exploited. However, this is not suitable for real-time place recognition deployment as
the entire test sequence is not known a priori.
Thus, to tackle this problem, we develop a static binning strategy by leveraging the \gls{vae} variance prediction, such that the uncertainty estimated is compared against a pre-defined prior distribution. 


\subsection{Uncertainty-aware Radar Localisation}

Although uncertainty estimation in radar place recognition tasks is still largely under-explored, there have been attempts to develop introspective radar systems for odometry. Adolfsson \textit{et al}~\cite{adolfsson2023tbv} present a radar system that verifies loop closure candidates introspectively by combining multiple place-recognition techniques, including tightly-coupled place similarity and odometry uncertainty search, creating loop descriptors from origin-shifted scans, and delaying loop selection until after verification. In contrast, our system does not rely on a secondary odometry estimation system, as uncertainty is inherent to the representation we learn for place recognition.
Aldera \textit{et al}~\cite{aldera2019could} focus only on odometry, and use inertial sensors to learn an uncertainty classifier, which we do not.

\section{Method}

In this section, we present the proposed modules for the variational uncertainty-aware radar place recognition method including metric space learning (\cref{Sec:VarCon}), mapping (\cref{Sec: MapMaintain}), and querying (\cref{Sec:IntroQuery}). We explain how these modules are integrated to form a unified system.

\begin{figure*}[!h]
\centering
\includegraphics[width=0.7\textwidth]{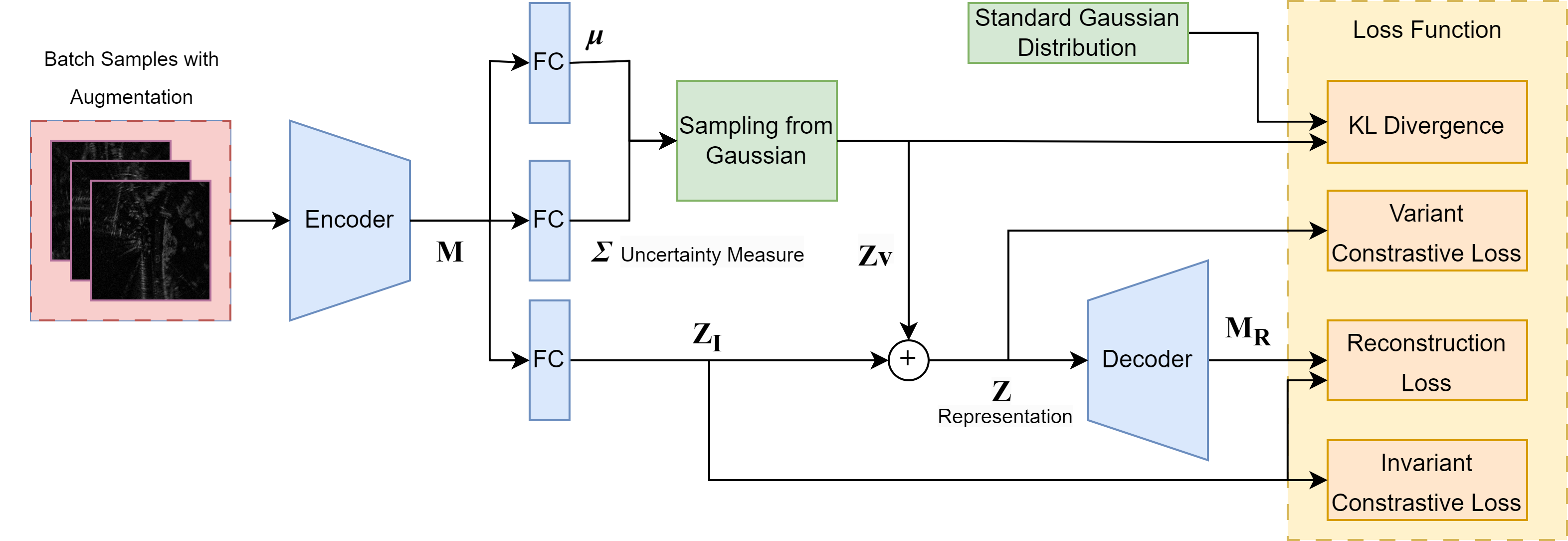}
\caption{
\textbf{Overview of Variational Contrastive Learning Framework}, based on~\cite{Lin2018dvml}.
A metric space is learnt through an encoder-decoder structure with two reparameterised parts: a deterministic embedding for recognition and a set of parameter modelling multivariate Gaussian distribution with its variance serving as the uncertainty measure. The overall learning is jointly driven by both reconstruction and contrastive loss to ensure an informative and discriminative hidden representation for radar scans.}
\label{fig:VariationalContrastiveFramework}
\vspace{-6mm}
\end{figure*}

\subsection{Variational Contrastive Learning} \label{Sec:VarCon}

The work in this section is both a key enabler of our core contributions, map maintenance and introspective querying -- see below, but is also a novel integration of deep variational metric learning (DVML)~\cite{Lin2018dvml} with radar place recognition~\cite{gadd2021icar,gadd2021icraws} and a new way to characterise uncertainty in place recognition.
Indeed, we show in~\cref{sec:results} that learning uncertainty in this way yields more calibrated introspection than in self-teaching uncertainty estimation (STUN)~\cite{cai2022stun}.


As illustrated in \cref{fig:VariationalContrastiveFramework}, we adopt a \gls{vae} structure to disentangle the radar scan embedding into a noise-induced variant part $Z_V$, which captures the variance of prediction-irrelevant uncertainty sources, and a semantic invariant part $Z_I$, for the essential features of the scene representation. The variant part is later sampled from a prior multivariate isotropic Gaussian distribution and added to the invariance part to form the overall representation $Z=Z_{I}+Z_{V}$. The variant output is directly used as the uncertainty measure.

The assumption is we only consider the aleatoric uncertainty in model prediction caused by the inherent ambiguity and randomness in data as the main source of uncertainty \cite{gal2016uncertainty}. In particular for radar scanning, this is likely due to speckle noise, saturation and temporary occlusion. The standard metric learning approaches, regardless of the loss function chosen, tend to enforce identical embeddings between positive sample pairs while ignoring the underlying variances between them. However, this can lead to the model being insensitive to minor features and overfitting to the training distribution. Thus, to model the noise variance, we use the extra probabilistic variance output in the \gls{vae} structure to estimate the aleatoric uncertainty. 

To establish such a noise-aware representation for radar perception, we use four loss functions to guide the overall training.
We choose to demonstrate the benefit of our specific method of learning uncertainty in the setting of losses which feature increased numbers of \textit{negative} examples -- this having been proven in STUN~\cite{cai2022stun} to perform best (i.e. quadruplet, with $2$ negatives).
In this area, the state-of-the-art in radar place recognition~\cite{gadd2021icar} already uses a loss with many (i.e. more than $2$) negative samples -- so, we extend upon this one.
For a fair comparison, we compare the uncertainty mechanism of STUN with ours under the same loss regime -- see further experimental details in~\cref{Sec:Baseline}.

\textbf{(1) Invariant contrastive loss}
on the deterministic representation $Z_I$ to disentangle the task-independent noise from radar semantics such that the invariant embedding contains sufficient causal information; and

\textbf{(2) Variant contrastive loss} between on the overall representation $Z$ to establish a meaningful metric space. Both contrastive losses take the following form in \cref{Eq:BCLoss}.
\begin{equation}
L_{Con}=-\sum_{i}^{m} \log P\left(i \mid \hat{\mathbf{x}}_i\right)-\sum_{i}^{m} \sum_{j \neq i}^{m} \log \left(1-P\left(i \mid \mathbf{x}_j\right)\right)
\label{Eq:BCLoss}
\end{equation}
where a batch consists of $m$ samples $\left\{\mathbf{x}_1, \mathbf{x}_2, \ldots, \mathbf{x}_m\right\}$ and synthetically rotated temporal proximal frames augmentations $\left\{\hat{\mathbf{x}}_1, \hat{\mathbf{x}}_2, \ldots, \hat{\mathbf{x}}_m\right\}$ using a ``spinning'' strategy \cite{gadd2021icar}, which is simply rotation augmentation, for rotation invariance.
We aim to maximise the probability that an augmented sample $\hat{x}_i$ is recognised as the original instance $x_i$ while minimising the probability of the reversed case. The probability of recognition is approximated as \cref{Eq:BCLoss_Prob}.
\begin{equation}
P\left(i \mid \mathbf{x}_j\right)=\frac{\exp \left(\frac{\mathbf{Z}_i^T \mathbf{Z}_j}{\tau}\right)}{\sum_{k=1}^m \exp \left(\frac{\mathbf{Z}_k^T \mathbf{Z}_j}{\tau}\right)}, j \neq i
\label{Eq:BCLoss_Prob}
\end{equation}
where the embeddings $\mathbf{z}$ are either $Z_I$ or $Z$ as in 1) and 2).

\textbf{(3) Kullback–Leibler (KL) divergence} between the learned Gaussian distribution and a standard isotropic multivariate Gaussian, which is our prior assumption on the data noise. This ensures an identical distribution across all the sample noises and provides a static reference for the absolute value of variant output.
\begin{equation}
D_{\mathrm{KL}}=\sum_{z \in \mathcal{Z_V}} \mathcal{N}\left( \boldsymbol{\mu}, \boldsymbol{\Sigma} \right) \log \left(\frac{\mathcal{N}\left( \boldsymbol{\mu}, \boldsymbol{\Sigma} \right)}{\mathcal{N}\left( \boldsymbol{0}, \boldsymbol{I} \right)}\right)
\label{Eq:KLDiv}
\end{equation}

\textbf{(4) Reconstruction loss} between the extract feature map $M$ and decoder output $M_R$, which forces the overall representation $Z$ to contain sufficient information in the original radar scans for reconstruction. However, instead of pixel-level radar scan reconstruction, we only reconstruct a lower dimensional feature map to reduce the computational cost in the decoding process. 
\begin{equation}
L_{Rec}=\left\lVert M_R-M\right \rVert_2
\label{Eq:ReconLoss}
\end{equation}
While the vanilla \gls{vae} structure driven by only KL divergence and reconstruction loss also provides latent variance, it is considered unreliable for uncertainty estimation  \cite{gaussianvae} due to its well-known problem of posterior collapse \cite{he2019lagging} and vanishing variance \cite{vanishing}.
We confirm this experimentally in~\cref{sec:results}.
Such ineffectiveness is mainly due to the imbalance of two losses during the training process: when the KL divergence dominates, the latent space posterior is forced to equal the prior whereas when the reconstruction loss dominates, the latent variance is pushed to zero. In our method, however, we achieve more stable training by introducing the variant contrastive losses as an extra regulariser, where the variance is driven to keep a robust boundary among clustering centres in metric space. Hence, we obtain a more reliable latent space variance reflecting the underlying aleatoric uncertainty of radar perception.
We show that this VAE-like uncertainty mechanism \textit{alongside} a metric learning objective is more well-calibrated to place recognition than STUN~\cite{cai2022stun}, vanilla VAEs~\cite{kingma2013auto}, MC Dropout~\cite{gal2016dropout}, etc, which we prove experimentally in~\cref{sec:results}.

\subsection{Continual Map Maintenance} \label{Sec: MapMaintain}

\begin{figure}[h]
\centering
\includegraphics[width=0.75\columnwidth]{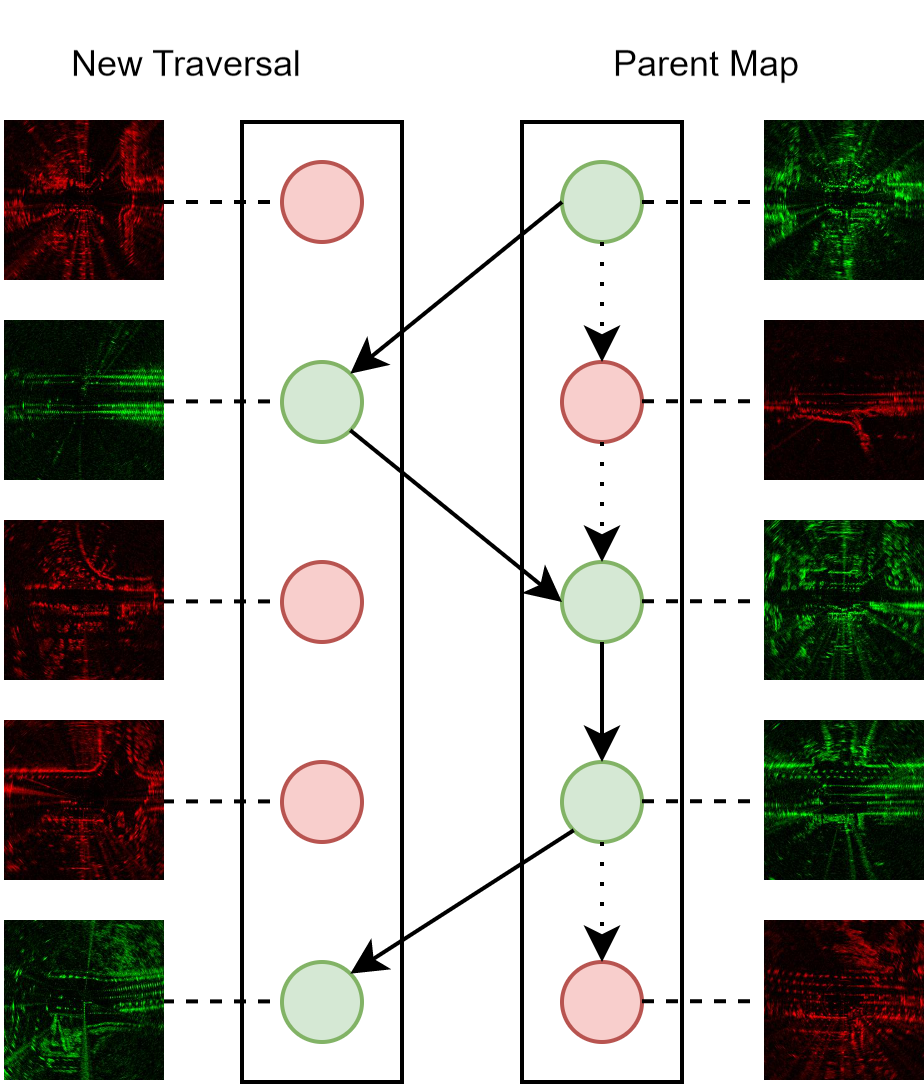}
\caption{\textbf{Illustration of Map Maintenance.} The red and green nodes each represent radar scans with higher and lower uncertainty respectively. We always maintain a parent map as the localisation reference consisting of only scans with the lowest uncertainty at each location. }
\label{fig:MapMaintenance}
\vspace{-6mm}
\end{figure}

Continual Map maintenance is an important function of the online system since we aim to fully exploit the scanning data obtained during a lifetime of autonomous vehicle operation and improve the map in a recursive manner. The process of merging a new radar scan into the parent map, consisting of scans from previous traversals, is illustrated in \cref{algo1_app}. Each radar scan is represented by both a hidden representation $\boldsymbol{\mu}$ and an uncertainty measure $U$. In the merging process, we search for matched positive scans for each new scan with topological distance under a threshold $D_t$. If the new scan has a lower uncertainty then it will be integrated into the parent map and replace the matched scan, otherwise, it will be discarded. 
\begin{algorithm}[H]
\caption{Map Maintenance}
\label{algo1_app}
\begin{algorithmic}[1]
\renewcommand{\algorithmicrequire}{\textbf{Input:}}
\renewcommand{\algorithmicensure}{\textbf{Output:}}
\REQUIRE Parent Map: $M = \{\left( \boldsymbol{\mu}_{i}, \boldsymbol{U}_{i} \right) | i\in[0, n] \}$, Map GPS $\{D_{i}\} | i\in[0, N]$ , New Scan: $\left( \boldsymbol{\mu}_{j}, \boldsymbol{U}_{j} \right)$, Scan GPS $D_{j}$.  
\ENSURE  Updated Parent Map: $\underline{M} = \{\left( \boldsymbol{\mu}_{i}, \boldsymbol{U}_{i} \right) | i\in[0, n] \}$
\\ \textit{Find Matched Scan Index}:
\STATE  $\{\Delta D_{i}| i\in[0, n] \}$ $\gets$ Sort$(\lVert D_{i}-D_{j} \rVert)$
\IF {$\Delta D_{i} < D_{t}$}
\STATE  $\{U_1, U_2, ... U_m\}$ $\gets$ GetUncertainty$(M,\Delta D_{i})$
\ENDIF
\\ \textit{Uncertainty Comparison and Replacement}:
\FOR {$i = 0$ to $m$}
\IF {($U_{j} > U_{i}$)}
\STATE $(\boldsymbol{\mu}_{i}, \boldsymbol{U}_{i}) \gets (\boldsymbol{\mu}_{j}, \boldsymbol{U}_{j})$
\ENDIF
\ENDFOR
\RETURN $\underline{M}$ 
\end{algorithmic}
\end{algorithm}
As shown in \cref{fig:MapMaintenance}, The replacement of higher uncertainty scans will remove noisy scans, which potentially lead to incorrect prediction, such as saturation and range uncertainty.

By iteratively performing the maintenance process for all obtained scans, we can gradually enhance the quality of the integrated parent map. Thus, the maintenance algorithm can serve as an effective online deployment strategy as it continually exploits multiple experiences of the same route traversal to boost recognition performance and, in the meanwhile, preserve a \textit{constant parent map size} resulting in budgeted computation and storage cost. 

In order to facilitate this map maintenance, we currently use GPS, but it is important to note that this step is purely offline (place recognition is performed entirely online, with radar alone) and that this offline step may be readily be replaced by whatever underlying map representation the place recognition queries are issued against, i.e. one built by radar-only mapping and motion estimation~\cite{cen2019radar,demartini2020kradar,adolfsson2023tbv,weston2022fastmbym}.
It is also important to note that our results in~\cref{sec:results} show that even without this map maintenance, our uncertainty-aware representation of~\cref{Sec:VarCon} outperforms previous methods, but that this method is capable of significantly boosting localisation performance by being careful about the learned uncertainty of map contents over many repeat traversals.


\subsection{Introspective Query Rejection} \label{Sec:IntroQuery}

During the inference, we use the Euclidean distance between the query and scanned map in a metric space as a measure of similarity. The top $k$ pairs with the highest similarity serve as the prediction result. To further enhance recognition performance and achieve introspective prediction, we also develop an uncertainty rejection mechanism.

\textbf{Static Thresholding}
Since the aleatoric uncertainty is measured against a standard Gaussian distribution $\mathcal{N}\left(\boldsymbol{0}, \boldsymbol{I} \right)$, the estimated variances in all dimensions are close to 1. 
Thus, we can use two hyperparameters $\Delta$ and $N$ to fully define the scale and resolution of uncertainty rejection respectively.
The resultant thresholds $T$ are defined as follows:
\begin{equation}
T = \{(1-\Delta) + n \times \frac{2\Delta}{N} | n \in [0, N]\}
\label{Eq:Threshold}
\end{equation}
Given a scan with $m$ dimension latent variance $\Sigma$, we average across all the dimensions and obtain a scalar uncertainty measure $U = \frac{1}{m} \sum_{i}^{m} \Sigma_{i}$. 

\textbf{Prediction Rejection}
At inference time, we perform introspective query rejection, where the query scans with a higher variance than a defined threshold will be rejected from recognition. Existing methods, such as STUN and MC Dropout, dynamically divide the uncertainty range of batch samples into threshold levels. However, this imposes the requirement of multiple samples during inference and can result in unstable rejection performance, particularly when a small number of samples are available. In contrast, our static thresholding strategy offers sample-independent threshold levels and provides consistent single-scan uncertainty estimation and rejection. This feature is crucial for the real-time deployment of the place recognition system, as the radar scan is acquired on a frame-by-frame basis during the driving process.



\section{Experimental Setup}
\subsection{Datasets}

We evaluate the performance using two standard benchmarks. (1) \textit{\textit{Oxford Radar RobotCar}} \cite{RadarRobotCarDatasetICRA2020}: 
It contains radar scans from 30 traversals of the same route around Oxford city centre equivalent to approximately \SI{300}{\kilo\meter} driving. We follow the train-test split of S{\u{a}}ftescu \textit{et al} \cite{suaftescu2020kidnapped}. 
An unseen route representing forward and backward traversal of both urban and vegetation environments is used for testing, and all the rest of the routes are used for training. We use five test sequences\footnote{\texttt{2019-01-18-14-46-59-radar-oxford-10k, 2019-01-18-15-20-12-radar-oxford-10k, 2019-01-18-14-14-42-radar-oxford-10k,
    2019-01-14-14-15-12-radar-oxford-10k,
    2019-01-16-14-15-33-radar-oxford-10k}} where two arbitrary sequences can form a mapping and localisation pair.
We report evaluation metric averages over all possible combinations\footnote{there are in total $20$ pairs, $10$ unique pairs of sequences, but the assignment of reference/query being important for introspective query rejection}. (2) \textit{MulRan} \cite{gskim-2020-mulran}: It contains radar scans from the traversal of different routes at four locations in Daejeon equivalent to around \SI{120}{\kilo\metre} driving. We train on all sequences from three locations, \texttt{DCC}, \texttt{KAIST}, and \texttt{Sejong}, then test on the same route in \texttt{Riverside} of two sequences, which alternate to serve for mapping and localisation in two testing trails. 

Both datasets encompass data collected from a CTS350-X Navtech \gls{fmcw} scanning radar. This particular radar system is devoid of Doppler information and is mounted on a platform in such a manner that its axis of rotation is perpendicular to the driving surface. The operating frequency of this radar system is situated within the range of \SIrange{76}{77}{\giga\hertz}, which enables it to generate up to \SI{3768}{} range readings with a resolution of \SI{4.38}{\centi\meter}. The total range of the radar system is \SI{165}{\metre}, and each range reading corresponds to one of the \SI{400}{} azimuth readings, which have a resolution of \SI{0.9}{\degree} degrees. The radar system is characterized by a scan rotation rate of \SI{4}{\hertz}.

\subsection{Baselines}\label{Sec:Baseline}

\textbf{Benchmarking} the recognition performance is done by comparison to several existing methods, including the vanilla \gls{vae} \cite{kingma2013auto}, the state-of-the-art radar place recognition method by Gadd \textit{et al}~\cite{gadd2021icar} (referred as BCRadar), and the non-learning-based method RingKey \cite{scancontext} (one part of ScanContext, without rotation refinement). Additionally, the performance is compared against MC Dropout \cite{gal2016dropout}, and STUN \cite{cai2022stun}, which serve as uncertainty-aware place recognition baselines.
We note that our testing of BCRadar uses a different list of dataset/sequence combinations than in~\cite{gadd2021icar}.

\textbf{Ablation study}
In order to evaluate the effectiveness of our proposed Introspective Query (Q) and Map Maintenance (M) modules, we conduct an ablation study by comparing different variants of our method, denoted as OURS(O/M/Q/QM), which are as follows
\begin{enumerate}
\item O: No map maintenance, no introspective querying
\item M: Map maintenance only
\item Q: Introspective querying only
\item QM: Both map maintenance and introspective querying
\end{enumerate}
Specifically, we compare the performance of O against M for recognition performance and Q against QM for uncertainty estimation performance. In each group of the variant comparison, we perform independent mapping inference over multiple localisation sequences and report the average metric, or merge multiple localisation sequences and report a single aggregate metric.


\textbf{Common settings}
To ensure a fair comparison, we adopt a common batch contrastive loss \cite{ye2019unsupervised} as for all metric learning-based methods, thus enabling a training regime with a consistent loss function across the benchmarking.

\subsection{Implementation details}
{\bf Scan settings}
For all methods, we transform polar radar scans of with $A = 400$ azimuths and $B = 3768$ bins of size \SI{4.38}{\centi\metre} to Cartesian scans with side-length $W = 256$ and bin size \SI{0.5}{\metre}.

{\bf Training hyperparameters}
We use a VGG-16 \cite{simonyan2014very} as the backbone feature extractor with a linear layer to project the extracted feature to a lower embedding dimension of $d=128$. We train all baselines for \num{10} epochs in \textit{\textit{Oxford Radar RobotCar}} and \num{15} epochs in \textit{MulRan} respectively\footnote{STUN is finetuned for another \num{5} epochs for knowledge distillation}, with a learning rate of $1\mathrm{e}{-5}$, batch size of \num{8}. In adapting the batch criterion of~\cite{ye2019unsupervised}, we use a negative margin in the embedding space of \num{0.1} and temperature $\tau$ of \num{1}. 

\subsection{Evaluation Metric}


To assess the place recognition performance, we use \texttt{Recall@N (R@N)} metric, which is the accuracy of the localisation by determining whether at least one candidate amongst $N$ candidates is close to the ground truth, as indicated by GPS/INS. This is particularly important to safety assurance in the autonomous driving application because it reflects the system's calibration to false negatives rate.
We also use \texttt{Average Precision (AP)} to measure the mean precision across all recall levels. Finally, we employ \texttt{F-scores} with $\beta=2/1/0.5$ to assign the increasing level of importance to recall over precision as an aggregate metric assessing overall recognition performance. 
For all metrics, we impose a boundary of \SI{25}{m} such that all radar scans within it are considered positive pairs to a mapping scan. Similarly, a \SI{50}{m} lower boundary defines the corresponding true negative pairs. 

Moreover, to assess the uncertainty estimation performance. We use \texttt{Recall@RR}, where we perform introspective query rejection and evaluate \texttt{Recall@N=1} over different uncertainty threshold levels -- rejecting all queries of scans with uncertainty greater than a threshold.
We thus reject between \SIrange{0}{100}{\percent} of queries.
A reliable uncertainty measure should reflect the confidence level of the model's predictions, and the rejection of low-confidence scans should result in improved recognition performance (i.e. we should not reject certain, good results). Our goal is to achieve higher improvement in recognition performance with a lower rate of rejections.

\section{Results}
\label{sec:results}

\subsection{Place Recognition Performance}

As demonstrated in \cref{tab1:oxfordcarPR} of the \textit{Oxford Radar RobotCar} experiments, our methods, utilizing only the metric-learning module, achieve the highest performance across all the metrics. Specifically, in terms of \texttt{Recall@1}, our approach OURS(O) showcases the efficacy of the variance-disentangled representation learned through the variational contrastive learning framework, resulting in superior 90.46\% recognition performance. This is further supported with the \textit{MulRan} experiment results demonstrated in \cref{tab2:MulranPR}, where our method outperforms all the other methods on \texttt{Recall@1} and aggregate \texttt{F-scores} and \texttt{AP}. Although \gls{vae} outperforms our method in \texttt{Recall@5/10} in the \textit{Mulran} experiment, the best \texttt{F-1/0.5/2} and \texttt{AP} of our method in both settings indicate an overall more accurate and robust recognition performance with high precision and recall.
Furthermore, \gls{vae} uncertainty is shown below (\cref{sec:results:uncertainty}) to be poor for introspective query rejection, while our learned uncertainty is useful in both settings.

Moreover, by further utilizing the Continual Map Maintenance in \textit{Oxford Radar RobotCar}, we are able to further improve the \texttt{Recall@1} to 93.67\%, outperforming the current state-of-the-art method, STUN, by a margin of 4.18\%. This further proves the effectiveness of learnt variance as a valid uncertainty measure and the strategy of uncertainty-based map integration in improving place recognition performance.

\begin{table}
\centering
\caption{Oxford Radar RobotCar Recognition Performance. The best and second-best results are bold and underlined, respectively.}
\label{tab1:oxfordcarPR}
\begin{tabular}{l|ccc|ccc|c} 
\toprule
 & \multicolumn{3}{c|}{R@N=1/5/10} & \multicolumn{3}{c|}{F-1/0.5/2}     & AP      \\ 
\hline
BCRadar &$85.39$ &$88.59$ &$90.52$ &$0.89$ &$0.83$ &$0.75$ &$0.95$\\
VAE &$48.35$ &$62.88$ &$69.81$ &$0.56$ &$0.48$ &$0.56$ &$0.70$\\
RingKey &$83.68$ &$86.49$ &$88.32$ &$0.85$ &$0.81$ &$0.71$ &$0.94$\\
STUN &$89.49$ &$92.23$ &$93.71$ &$0.92$ &$0.87$ &$0.76$ &$0.98$\\
MC Dropout &$88.51$ &$91.57$ &$93.24$ &$0.91$ &$0.85$ &$0.76$ &$0.97$\\
\hline
OURS(O) &$\underline{90.46}$ &$\underline{93.76}$ &$\underline{95.22}$ &$\underline{0.93}$ &$\underline{0.87}$ &$\underline{0.77}$ &$\underline{0.98}$\\
OURS(M) &$\textbf{93.67}$ &$\textbf{96.24}$ &$\textbf{97.39}$ &$\textbf{0.95}$ &$\textbf{0.91}$ &$\textbf{0.78}$ &$\textbf{0.99}$\\
\bottomrule
\end{tabular}
\vspace{-5mm}
\end{table}
\begin{table}
\centering
\caption{MulRan Recognition Performance. Results reported following the same practice as \cref{tab1:oxfordcarPR}.}
\label{tab2:MulranPR}
\begin{tabular}{l|ccc|ccc|c} 
\toprule
  & \multicolumn{3}{c|}{R@N=1/5/10} & \multicolumn{3}{c|}{F-1/0.5/2}      & AP      \\ 
  \hline
BCRadar &$32.23$ &$37.81$ &$40.70$ &$0.36$ &$0.31$ &$0.36$ &$0.43$\\
VAE &$\underline{36.47}$ &$\textbf{46.54}$ &$\textbf{52.58}$ &$\underline{0.41}$ &$\underline{0.35}$ &$\underline{0.42}$ &$\underline{0.50}$\\
RingKey &$27.53$ &$29.34$ &$30.17$ &$0.30$ &$0.27$ &$0.30$ &$0.36$\\
MC Dropout &$32.90$ &$37.50$ &$40.50$ &$0.35$ &$0.32$ &$0.34$ &$0.43$\\
STUN &$26.76$ &$31.15$ &$34.04$ &$0.31$ &$0.26$ &$0.31$ &$0.36$\\
\hline
OURS(O) & $\textbf{37.50}$ &$\underline{43.54}$ &$\underline{47.57}$ &$\textbf{0.44}$ &$\textbf{0.37}$ &$\textbf{0.43}$ &$\textbf{0.51}$\\

\bottomrule
\end{tabular}
\vspace{-5mm}
\end{table}

\subsection{Uncertainty Estimation Performance}
\label{sec:results:uncertainty}

\begin{figure}[h]
\includegraphics[width=0.925\linewidth]{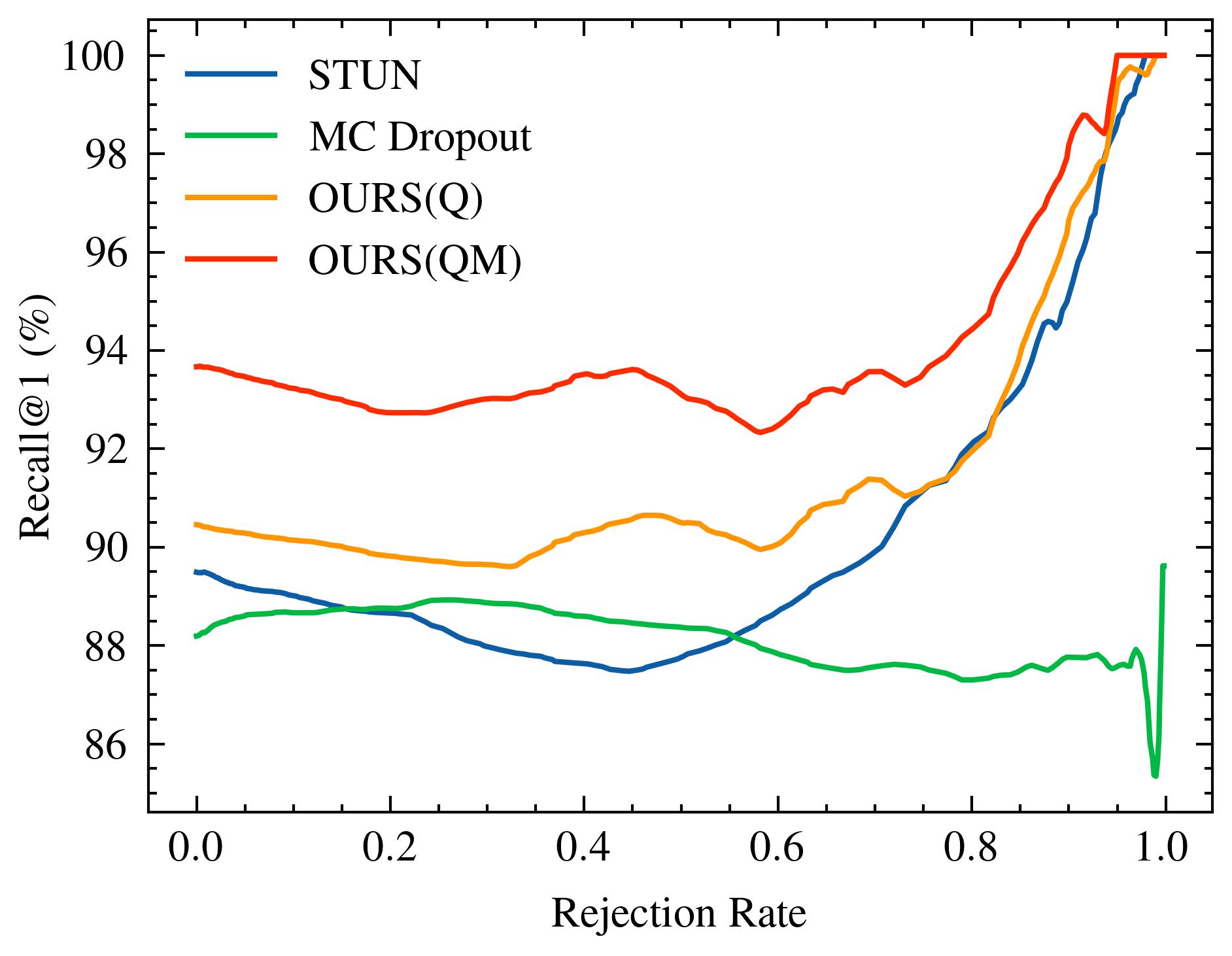}
\caption{\textit{Oxford Radar RobotCar} introspective query rejection performance. \texttt{Recall@1} increases/decreases as the percentage of uncertain query rejected increases. VAE's performance is not visualised as it is was much lower in comparison to other methods (specifically (48.42/48.08/18.48)\% for \texttt{Recall@RR=0.1/0.5/1.0}).}
\vspace{-2mm}
\label{fig:ox_rej}
\end{figure}

\begin{figure}[h]
\includegraphics[width=0.925\linewidth]{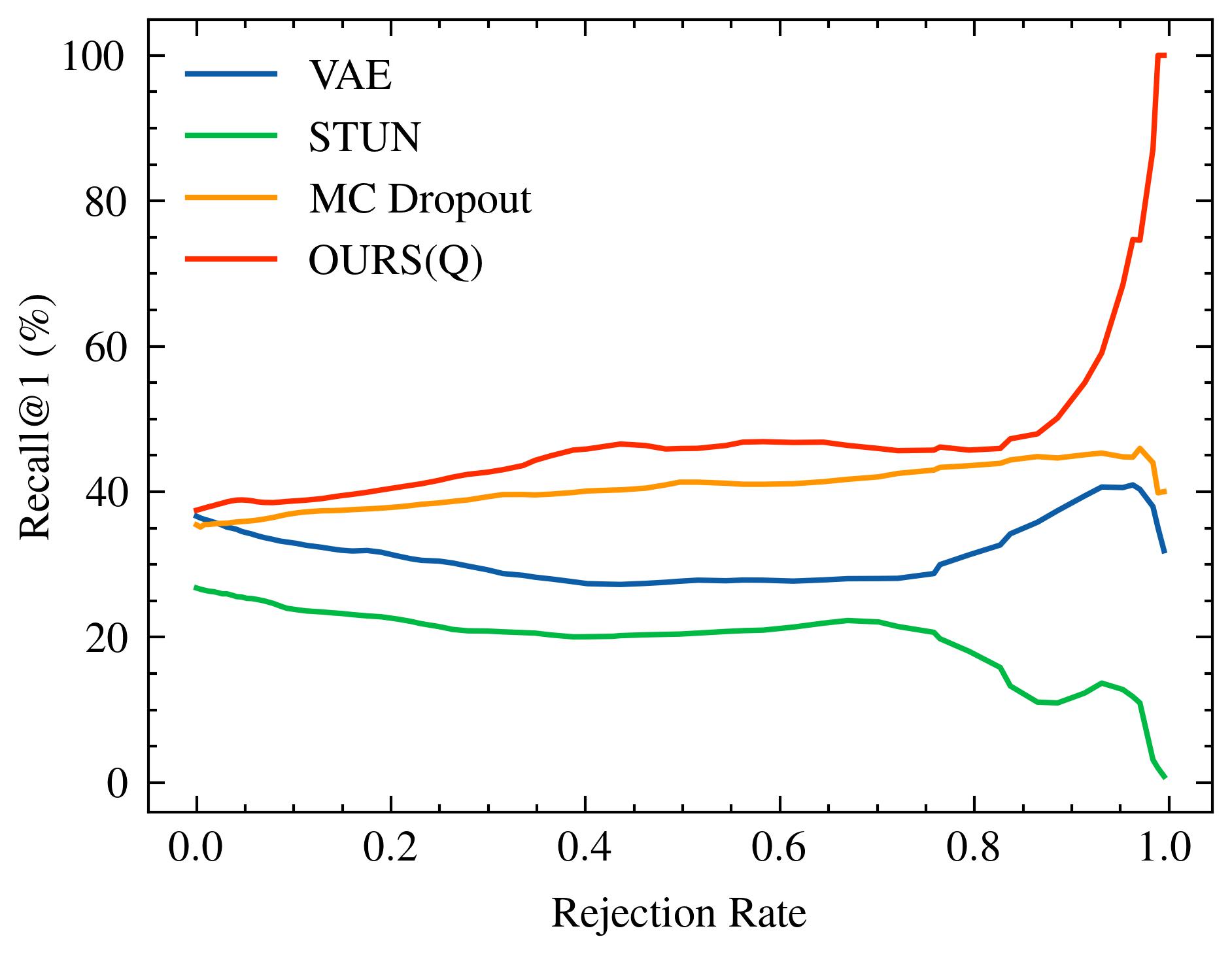}
\caption{Mulran introspective query rejection performance reported the same format as in \cref{fig:ox_rej}. \vspace{-6mm}}
\label{fig:mulran_rej}
\end{figure}

The change in recognition performance, specifically the \texttt{Recall@1}, with an increasing percentage of rejected uncertain queries, is illustrated in \cref{fig:ox_rej} for the \textit{Oxford Radar RobotCar} experiment and in \cref{fig:mulran_rej} for the \textit{MulRan} experiment.
Notably, our method is the only one to exhibit a consistent improvement in recognition performance as the rejection rate of uncertain queries increases in both experimental settings -- \textit{Oxford Radar RobotCar} and \textit{MulRan} -- whereas STUN increases for \textit{Oxford Radar RobotCar} but not  \textit{MulRan} and MC Dropout increases for \textit{MulRan} but not \textit{Oxford Radar RobotCar}.
For \textit{Oxford Radar RobotCar} in~\cref{fig:ox_rej}, it is interesting to note that Ours(QM) and Ours(Q) but outperform the initial \texttt{Recall@1} at a lower rejection rate than STUN does -- about \SI{40}{\percent} for Ours(QM) and Ours(Q) but a much more comprehensively rejection of \SI{70}{\percent} than for STUN.
This result serves as evidence of the robustness and effectiveness of our uncertainty measure.

In the \textit{MulRan} experiments, OURS(Q) is the only method that consistently and smoothly improves the \texttt{Recall@RR} metric as the rejection rate increases. Compared to \gls{vae} and STUN, which also estimate aleatoric uncertainty like our method, OURS(Q) achieves an improvement of +(1.32/3.02/8.46)\% on \texttt{Recall@RR=0.1/0.2/0.5}, while \gls{vae} and STUN show a decline of -(3.79/5.24/8.80)\% and -(2.97/4.16/6.30)\%, respectively. Although in the \textit{Oxford Radar RobotCar} experiments, all methods initially experience fluctuations in performance as the rejection rate increases, ultimately, OURS(Q/QM) outperforms both \gls{vae} and STUN with a higher \texttt{Recall@RR} metric at all rejection rates. The superior performance of OURS over \gls{vae} suggests the effectiveness of using a regularizing invariant discriminative loss to address the issues of posterior collapse and vanishing variance in the vanilla \gls{vae} structure. Furthermore, the better performance of OURS compared to STUN indicates that our variational contrastive learning framework can be more effective than knowledge distillation in learning a reliable variance for uncertainty estimation.

On the other hand, compared with MC Dropout, which estimates the epistemic uncertainty resultant from biased data and model misspecification \cite{gal2016uncertainty}, although it has a higher increase in \texttt{Recall@1} at the early stage of rejection in \textit{Oxford Radar RobotCar} experiment, its performance is generally lower than ours and fails to achieve greater improvement as the rejection rate increases further. Also, in \textit{MulRan} experiments, it fails to produce reliable uncertainty estimates, as indicated by the general decreasing trend in \texttt{Recall@RR}. These results suggest that aleatoric uncertainty plays a more significant role in causing mispredictions in place recognition than epistemic uncertainty.

Finally, comparing OURS(Q) and OURS(QM) in \textit{Oxford Radar RobotCar} experiment, we observe a similar change in \texttt{Recall@RR} pattern while a considerable gap exists between them. This suggests that the Introspective Query and Map Maintenance mechanisms independently contribute to the place recognition system and that each mechanism exploits the uncertainty measure in an indispensable way.



\section{Discussion}

\subsection{Qualitative Analysis and Visualisation}

\begin{figure}[h]
\centering
\includegraphics[width=0.95\linewidth]{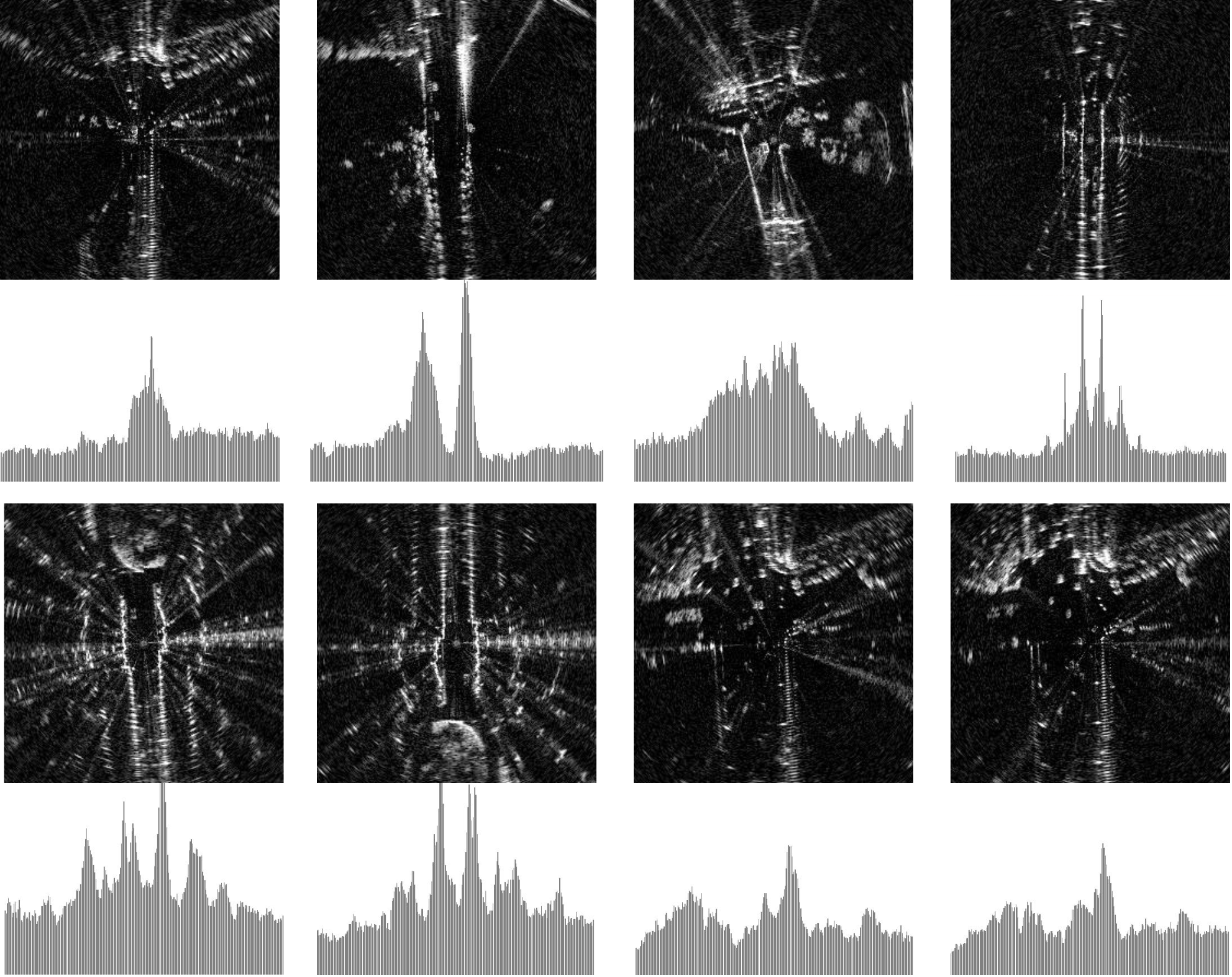}
\caption{\textbf{Visualisation of Radar Scans with Different Levels of Uncertainty.} 
The four examples on the left are from the \textit{Oxford Radar RobotCar Dataset} while the four examples on the right are from \textit{MulRan}.
We show samples with the Top-10 highest (top) / lowest (bottom) uncertainty. The radar scan is displayed in a Cartesian coordinate with enhanced contrast. The histogram below each image shows the Ring Key descriptor \cite{scancontext} extracted feature of the intensity across all the azimuths. \vspace{-2.5mm}}
\label{fig:vis_uncertainty}
\end{figure}

To qualitatively assess the source of uncertainty in radar perception, we provide a visual comparison of the high/low uncertainty samples from both datasets estimated with our method. As shown in \cref{fig:vis_uncertainty}, the high-uncertainty radar scans usually demonstrate heavy motion blur and sparse undetected regions, while the low-uncertainty scans usually contain distinct features with a stronger intensity across the histogram. 

This further supports our hypothesis on the source of uncertainty in radar perception and serves as qualitative evidence of the effectiveness of our uncertainty measure that captures this data noise.

\subsection{Dataset Difficulty: Uncertain Environments}
In our benchmark experiments, we observed a considerable discrepancy in the recognition performance between the two datasets. 
The difficulty of \textit{MulRan} as compared to \textit{Oxford Radar RobotCar} for both place recognition and metric localisation is already well-documented in~\cite{tang2021self,gadd2021icar}.
We posit that the scale of available training data can be a plausible cause. The training set in \textit{Oxford Radar RobotCar} comprises over \SI{300}{\kilo\metre} driving experiences, while the \textit{MulRan} dataset includes only around \SI{120}{\kilo\metre}. 
However, also consider
the drop in performance for the \textit{non-learned} Ring Key descriptor method. This suggests potential inherent indistinguishable features in radar scene perception. For instance, we found that environments with sparse open areas usually lead to identical scan and suboptimal recognition performance.
We include results on this dataset to exhibit what happens to our system and the various baselines in these cases of high uncertainty.
As shown in~\cref{fig:mulran_rej}, our learned uncertainty is the most useful in these difficult scenarios.


\section{Conclusion}
We have presented a novel application of uncertainty estimation to learned radar place recognition.
We bolster the performance of an invariant instance feature learning approach with a noise-aware latent representation with variational learning. 
We apply this to
(1) continual map maintenance for online system deployment in long-term autonomy, where reference sequences are constructed from low-uncertainty scans only; and (2)
introspective query rejection with static thresholding for sequential single-scan real-time inference, where our uncertainty gives the system the ability to understand when it is likely to be incorrect
In tests over two sizeable public radar datasets, we outperform the previous state-of-the-art in radar place recognition, and other uncertainty estimation techniques.


\section*{Acknowledgements}

This work was supported by EPSRC Programme Grant ``From Sensing to Collaboration'' (EP/V000748/1).
We would also like to thank our partners at Navtech Radar.

{\small
\bibliographystyle{ieeetr}
\bibliography{main}

\begin{thebibliography}{10}

\bibitem{domainbed}
I.~Gulrajani and D.~Lopez-Paz, ``In search of lost domain generalization,''
  {\em arXiv preprint arXiv:2007.01434}, 2020.

\bibitem{checchinscan}
P.~Checchin, F.~Gérossier, C.~Blanc, R.~Chapuis, and L.~Trassoudaine, ``Radar
  scan matching slam using the fourier-mellin transform,'' pp.~151--161, 2009.

\bibitem{cen2019radar}
S.~H. Cen and P.~Newman, ``Radar-only ego-motion estimation in difficult
  settings via graph matching,'' in {\em 2019 International Conference on
  Robotics and Automation (ICRA)}, pp.~298--304, IEEE, 2019.

\bibitem{hong2020radarslam}
Z.~Hong, Y.~Petillot, and S.~Wang, ``Radarslam: Radar based large-scale slam in
  all weathers,'' in {\em 2020 IEEE/RSJ International Conference on Intelligent
  Robots and Systems (IROS)}, pp.~5164--5170, IEEE, 2020.

\bibitem{scancontext}
G.~Kim and A.~Kim, ``Scan context: Egocentric spatial descriptor for place
  recognition within 3d point cloud map,'' in {\em 2018 IEEE/RSJ International
  Conference on Intelligent Robots and Systems (IROS)}, pp.~4802--4809, 2018.

\bibitem{gskim-2020-mulran}
G.~Kim, Y.~S. Park, Y.~Cho, J.~Jeong, and A.~Kim, ``Mulran: Multimodal range
  dataset for urban place recognition,'' in {\em Proceedings of the IEEE
  International Conference on Robotics and Automation (ICRA)}, 2020.

\bibitem{suaftescu2020kidnapped}
{\c{S}}.~S{\u{a}}ftescu, M.~Gadd, D.~De~Martini, D.~Barnes, and P.~Newman,
  ``Kidnapped radar: Topological radar localisation using
  rotationally-invariant metric learning,'' in {\em 2020 IEEE International
  Conference on Robotics and Automation (ICRA)}, pp.~4358--4364, IEEE, 2020.

\bibitem{gadd2021icar}
M.~Gadd, D.~De~Martini, and P.~Newman, ``{Contrastive Learning for Unsupervised
  Radar Place Recognition},'' in {\em Proceedings of the IEEE International
  Conference on Advanced Robotics (ICAR)}, 2021.

\bibitem{komorowski2021large}
J.~Komorowski, M.~Wysoczanska, and T.~Trzcinski, ``Large-scale topological
  radar localization using learned descriptors,'' in {\em Neural Information
  Processing: 28th International Conference, ICONIP 2021, Sanur, Bali,
  Indonesia, December 8--12, 2021, Proceedings, Part II 28}, pp.~451--462,
  Springer, 2021.

\bibitem{gadd2020look}
M.~Gadd, D.~De~Martini, and P.~Newman, ``Look around you: Sequence-based radar
  place recognition with learned rotational invariance,'' in {\em 2020 IEEE/ION
  Position, Location and Navigation Symposium (PLANS)}, pp.~270--276, IEEE,
  2020.

\bibitem{wang2021radarloc}
W.~Wang, P.~P. de~Gusmo, B.~Yang, A.~Markham, and N.~Trigoni, ``{RadarLoc:
  Learning to Relocalize in FMCW Radar},'' in {\em IEEE International
  Conference on Robotics and Automation (ICRA)}, 2021.

\bibitem{demartini2020kradar}
D.~De~Martini, M.~Gadd, and P.~Newman, ``{kRadar++: Coarse-to-fine FMCW
  Scanning Radar Localisation},'' {\em Sensors}, vol.~20, no.~21, p.~6002,
  2020.

\bibitem{cait2022autoplace}
K.~Cait, B.~Wang, and C.~X. Lu, ``Autoplace: Robust place recognition with
  single-chip automotive radar,'' in {\em 2022 International Conference on
  Robotics and Automation (ICRA)}, pp.~2222--2228, IEEE, 2022.

\bibitem{gadd2021icraws}
M.~Gadd, D.~De~Martini, and P.~Newman, ``{Unsupervised Place Recognition with
  Deep Embedding Learning over Radar Videos},'' {\em arXiv preprint
  arXiv:2106.06703}, 2021.

\bibitem{yin2021radar}
H.~Yin, X.~Xu, Y.~Wang, and R.~Xiong, ``Radar-to-lidar: Heterogeneous place
  recognition via joint learning,'' {\em Frontiers in Robotics and AI}, vol.~8,
  p.~661199, 2021.

\bibitem{gal2016dropout}
Y.~Gal and Z.~Ghahramani, ``Dropout as a bayesian approximation: Representing
  model uncertainty in deep learning,'' in {\em international conference on
  machine learning}, pp.~1050--1059, PMLR, 2016.

\bibitem{cai2022stun}
K.~Cai, C.~X. Lu, and X.~Huang, ``{STUN}: Self-teaching uncertainty estimation
  for place recognition,'' pp.~6614--6621, 2022.

\bibitem{warburg2021bayesian}
F.~Warburg, M.~J{\o}rgensen, J.~Civera, and S.~Hauberg, ``Bayesian triplet
  loss: Uncertainty quantification in image retrieval,'' in {\em Proceedings of
  the IEEE/CVF International Conference on Computer Vision}, pp.~12158--12168,
  2021.

\bibitem{shi2019probabilistic}
Y.~Shi and A.~K. Jain, ``Probabilistic face embeddings,'' in {\em Proceedings
  of the IEEE/CVF International Conference on Computer Vision}, pp.~6902--6911,
  2019.

\bibitem{kingma2013auto}
D.~P. Kingma and M.~Welling, ``Auto-encoding variational bayes,'' {\em arXiv
  preprint arXiv:1312.6114}, 2013.

\bibitem{vaepr1}
J.~Oh and G.~Eoh, ``Variational bayesian approach to condition-invariant
  feature extraction for visual place recognition,'' {\em Applied Sciences},
  vol.~11, 2021.

\bibitem{vaepr2}
H.~Wu and M.~Flierl, ``Learning product codebooks using vector-quantized
  autoencoders for image retrieval,'' in {\em 2019 IEEE Global Conference on
  Signal and Information Processing (GlobalSIP)}, pp.~1--5, 2019.

\bibitem{rey2021contentbased}
L.~A.~P. Rey, D.~Jarnikov, and M.~Holenderski, ``Content-based image retrieval
  from weakly-supervised disentangled representations,'' in {\em NeurIPS 2021
  Workshop on Deep Generative Models and Downstream Applications}, 2021.

\bibitem{8967633}
Y.~S. Park, J.~Kim, and A.~Kim, ``Radar localization and mapping for indoor
  disaster environments via multi-modal registration to prior lidar map,'' in
  {\em 2019 IEEE/RSJ International Conference on Intelligent Robots and Systems
  (IROS)}, pp.~1307--1314, 2019.

\bibitem{Lin2018dvml}
X.~Lin, Y.~Duan, Q.~Dong, J.~Lu, and J.~Zhou, ``Deep variational metric
  learning,'' in {\em European Conference on Computer Vision}, 2018.

\bibitem{burnett2021radar}
K.~Burnett, D.~J. Yoon, A.~P. Schoellig, and T.~D. Barfoot, ``Radar odometry
  combining probabilistic estimation and unsupervised feature learning,'' {\em
  arXiv preprint arXiv:2105.14152}, 2021.

\bibitem{taha2019unsupervised}
A.~Taha, Y.-T. Chen, T.~Misu, A.~Shrivastava, and L.~Davis, ``Unsupervised data
  uncertainty learning in visual retrieval systems,'' {\em arXiv preprint
  arXiv:1902.02586}, 2019.

\bibitem{adolfsson2023tbv}
D.~Adolfsson, M.~Karlsson, V.~Kubelka, M.~Magnusson, and H.~Andreasson, ``Tbv
  radar slam -- trust but verify loop candidates,'' {\em arXiv preprint
  arXiv:2301.04397}, 2023.

\bibitem{aldera2019could}
R.~Aldera, D.~De~Martini, M.~Gadd, and P.~Newman, ``What could go wrong?
  introspective radar odometry in challenging environments,'' in {\em 2019 IEEE
  Intelligent Transportation Systems Conference (ITSC)}, pp.~2835--2842, IEEE,
  2019.

\bibitem{gal2016uncertainty}
Y.~Gal, {\em Uncertainty in deep learning}.
\newblock PhD thesis, University of Cambridge.

\bibitem{gaussianvae}
J.~B{\"u}tepage, L.~Maystre, and M.~Lalmas, ``Gaussian process encoders: Vaes
  with reliable latent-space uncertainty,'' in {\em Machine Learning and
  Knowledge Discovery in Databases. Research Track}, pp.~84--99, Springer
  International Publishing, 2021.

\bibitem{he2019lagging}
J.~He, D.~Spokoyny, G.~Neubig, and T.~Berg-Kirkpatrick, ``Lagging inference
  networks and posterior collapse in variational autoencoders,'' {\em arXiv
  preprint arXiv:1901.05534}, 2019.

\bibitem{vanishing}
H.~Akrami, A.~Joshi, S.~Aydore, and R.~Leahy, ``Quantile regression for
  uncertainty estimation in vaes with applications to brain lesion detection,''
  in {\em Information Processing in Medical Imaging} (A.~Feragen, S.~Sommer,
  J.~Schnabel, and M.~Nielsen, eds.), pp.~689--700, 2021.

\bibitem{weston2022fastmbym}
R.~Weston, M.~Gadd, D.~De~Martini, P.~Newman, and I.~Posner, ``{Fast-MbyM:
  Leveraging Translational Invariance of the Fourier Transform for Efficient
  and Accurate Radar Odometry},'' in {\em International Conference on Robotics
  and Automation (ICRA)}, pp.~2186–--2192, May 2022.

\bibitem{RadarRobotCarDatasetICRA2020}
D.~Barnes, M.~Gadd, P.~Murcutt, P.~Newman, and I.~Posner, ``The oxford radar
  robotcar dataset: A radar extension to the oxford robotcar dataset,'' in {\em
  Proceedings of the IEEE International Conference on Robotics and Automation
  (ICRA)}, (Paris), 2020.

\bibitem{ye2019unsupervised}
M.~Ye, X.~Zhang, P.~C. Yuen, and S.-F. Chang, ``Unsupervised embedding learning
  via invariant and spreading instance feature,'' in {\em Proceedings of the
  IEEE/CVF Conference on Computer Vision and Pattern Recognition},
  pp.~6210--6219, 2019.

\bibitem{simonyan2014very}
K.~Simonyan and A.~Zisserman, ``Very deep convolutional networks for
  large-scale image recognition,'' {\em arXiv preprint arXiv:1409.1556}, 2014.

\bibitem{tang2021self}
T.~Y. Tang, D.~De~Martini, S.~Wu, and P.~Newman, ``Self-supervised learning for
  using overhead imagery as maps in outdoor range sensor localization,'' {\em
  The International Journal of Robotics Research}, vol.~40, no.~12-14,
  pp.~1488--1509, 2021.

\end{thebibliography}
}

\end{document}